\documentclass{article}
\usepackage{spconf,amsmath,graphicx,hyperref}
\usepackage{amsthm, amssymb, amsfonts, latexsym, mathtools, bm}
\usepackage{cite, booktabs, multirow}
\usepackage{tikz}
\usetikzlibrary{shapes.geometric, arrows.meta, positioning, calc, decorations.pathreplacing, decorations.pathmorphing}
\usepackage[table]{xcolor}
\definecolor{highlightgray}{gray}{0.90} 
\title{LIPSSM: STRUCTURALLY LIPSCHITZ-BOUNDED CASCADED STATE-SPACE MODEL \\ VIA METRIC TRANSFER BETWEEN CONSECUTIVE SSM LAYERS}
\name{Natsuki Yoshino\qquad Ren Uchida\qquad  Kazuki Matsumoto\qquad  Kohei Yatabe\thanks{This work was supported by JST FOREST Program (JPMJFR2330).}}
\address{Tokyo University of Agriculture and Technology (TUAT), Tokyo, Japan\vspace{-3pt}}

\newcommand{\transp}{\mathsf{T}}

\def\vecu{{\mathbf{u}}}
\def\vecx{{\mathbf{x}}}
\def\vecy{{\mathbf{y}}}

\def\bA{{\mathbf{A}}}
\def\bB{{\mathbf{B}}}
\def\bC{{\mathbf{C}}}
\def\bD{{\mathbf{D}}}

\def\bI{{\mathbf{I}}}
\def\bJ{{\mathbf{J}}}
\def\bK{{\mathbf{K}}}

\def\bM{{\mathbf{M}}}
\def\bP{{\mathbf{P}}}
\def\bQ{{\mathbf{Q}}}
\def\bR{{\mathbf{R}}}
\def\bS{{\mathbf{S}}}
\def\bY{{\mathbf{Y}}}
\def\bW{{\mathbf{W}}}
\def\bV{{\mathbf{V}}}

\def\bPhi{{\bm{\Phi}}}
\def\bPsi{{\bm{\Psi}}}
\def\bPi{{\bm{\Pi}}}

\def\NN{\mathcal{N}}
\def\LL{\mathcal{L}}
\def\Lip{\operatorname{Lip}}

\newtheorem{thm}{Theorem}
\newtheorem{proposition}[thm]{Proposition}
\newtheorem{lemma}[thm]{Lemma}
\newtheorem{remark}[thm]{Remark}

\newtheorem{dfn}[thm]{Definition}
\begin{document}
\ninept
\maketitle
\begin{abstract}
Lipschitz continuity is a fundamental principle in the design of certifiably robust deep neural networks (DNNs), wherein adjusting the Lipschitz constant, which quantifies network robustness, is of central theoretical importance. 
A standard approach to enforcing Lipschitz continuity requires each layer of a DNN to be Lipschitz continuous, thereby guaranteeing overall Lipschitz continuity. 
However, this layer-wise approach typically imposes overly conservative restrictions by producing a loose estimate of the overall Lipschitz constant, which limits the expressive capacity of the DNN and degrades empirical performance at a prescribed level of robustness.
To overcome this loose estimation, the recently proposed LipKernel transfers information across layers to yield a much tighter overall Lipschitz bound than conventional layer-wise construction. 
In this paper, we extend this concept to cascaded state-space models (SSMs) to construct Lipschitz-continuous DNNs capable of modeling longer-term dependencies. The proposed architecture, named LipSSM, is theoretically justified and empirically evaluated.
\end{abstract}


\section{Introduction}\label{sec:intro}
Ensuring input-output stability in deep neural networks (DNNs) is crucial in  machine learning and signal processing.
To establish such theoretical guarantees, the Lipschitz continuity of DNNs serves as a cornerstone. 
Formally, a DNN $\NN$ is $\rho$-Lipschitz continuous if $\Vert{}\NN({\vecu
}^{\mathrm{in}})-\NN(\widetilde{\vecu}^{\mathrm{in}})\Vert{} \le \rho \Vert{}\vecu^{\mathrm{in}}-\widetilde{\vecu}^{\mathrm{in}}\Vert{}$ for all admissible inputs $\vecu^{\mathrm{in}}$ and $\widetilde{\vecu}^{\mathrm{in}}$. 
The Lipschitz constant of $\NN$, denoted by $\Lip(\NN)$, is defined as the smallest $\rho$ satisfying this inequality.
Bounding the Lipschitz constant (e.g., $\Lip(\NN)\leq 1$) is essential for certified adversarial robustness \cite{cisse2017, tsuzuku2018, fazlyab19IQC}, as well as for stable model-based signal processing incorporating pretrained DNNs \cite{pnp2013, pnp2022zhang, lipsam_icassp, lipsam26full, pnp_converge2017}.
In a broader context, controlling the Lipschitz constant is crucial for improving generalization capabilities \cite{ulrike, bartlett}. 

Consider an $L$-layer DNN $\NN$ given by
\begin{equation}
\NN = \LL^{(L)} \circ \LL^{(L-1)} \circ \cdots \circ \LL^{(2)} \circ \LL^{(1)}. \label{eq:NN}
\end{equation}
To ensure $\Lip(\NN) \le \rho$,
typical approaches impose layer-wise bounds satisfying $\prod_{l=1}^{L}\Lip(\LL^{(l)}) \le \rho$ based on the property $\Lip(\NN) \le \prod_{l=1}^{L}\Lip(\LL^{(l)})$ 
\cite{1LipLayers, trockman21orthogonal, AOL22, meunier2022,wang23sandwich}.
However, this upper bound is often loose, and hence a constraint much more restrictive than the target bound $\rho$ is implicitly enforced on $\Lip(\NN)$, which in turn degrades the representational capability of $\NN$.

To address this issue, recent research has shifted focus from layer-wise constraints 
$\prod_{l=1}^{L} \Lip(\LL^{(l)}) \leq \rho$ toward directly imposing network-level constraints $\Lip(\NN) \leq \rho$.
Building upon control theory \cite{koelwijin21,verhoek23,pauli2024a} and state-space representations of convolution \cite{rosser1975, gramlich2026}, the parameterization scheme proposed by Pauli et al.\ \cite{pauli2023cayley} and its 2D extension, LipKernel \cite{LipKernel}, provide a structural Lipschitz guarantee for convolutional neural networks (CNNs) without sacrificing model expressivity. 
This framework utilizes state-space models (SSMs) \cite{fu2022hippo, gu22, gu24mamba} restricted to finite-time systems. 
Since SSMs possess the capability to model long-term dependencies in sequential data, extending the finite-time LipKernel concept to more general SSMs with provable Lipschitz bounds offers a promising direction for certifiably robust deep learning.

\begin{figure}
    \centering
    \includegraphics[width=0.9\linewidth]{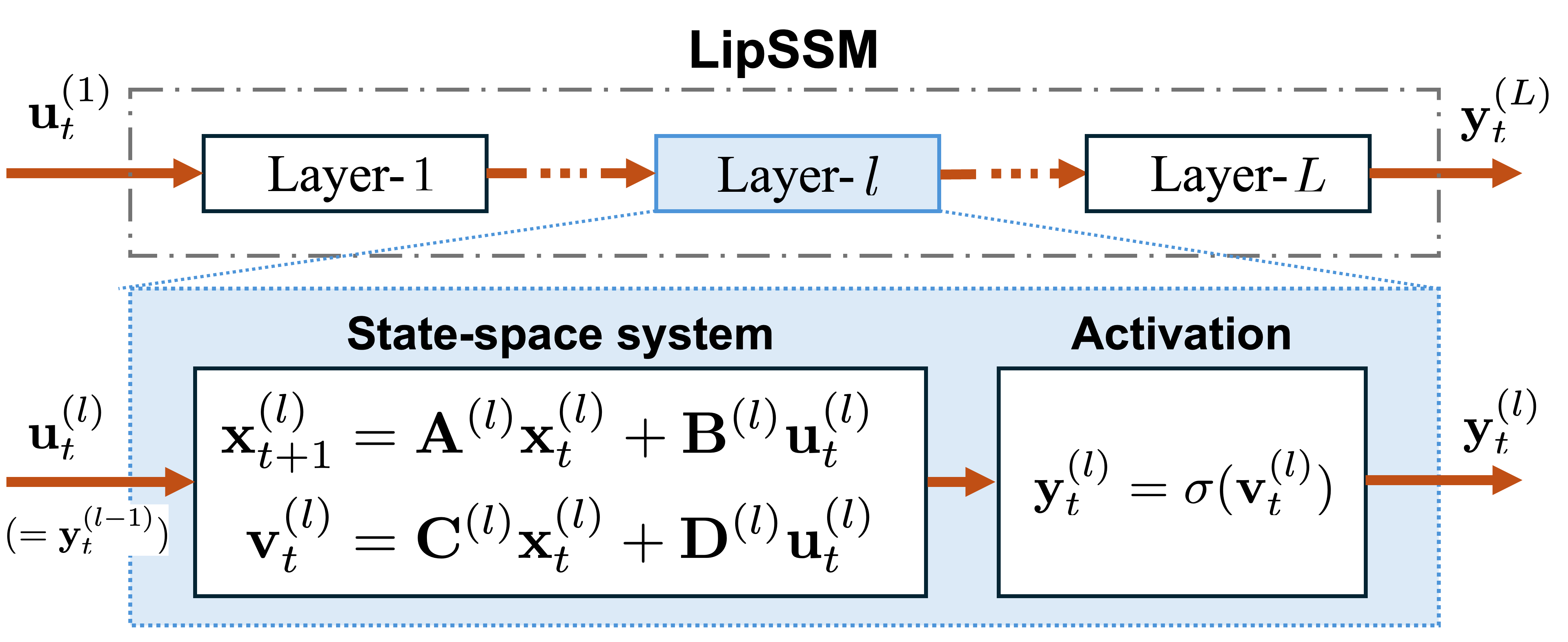}
    \vspace{-6pt}
    \caption{Block diagram of proposed LipSSM.}
    \label{fig:concept}
\end{figure}

In this paper, we propose LipSSM, a state-space network architecture (illustrated in Fig.\ \ref{fig:concept}) that structurally enforces a prescribed Lipschitz bound. 
Unlike conventional approaches to Lipschitz-continuous recurrent neural networks (RNNs) that rely on layer-wise constraints \cite{helfrich2018,L2RU25, R2DN26, martinelli23REN, REN24}, the proposed method transfers information across layers to achieve tighter estimates, thereby enhancing expressive power. 
Our main contributions are threefold: 
(i) formulating a novel parameterization of system matrices in SSMs to enforce Lipschitz continuity;
(ii) mathematically establishing that LipSSM achieves network-level Lipschitz certification by construction; 
and (iii) empirically validating our theoretical guarantees while demonstrating the efficacy of LipSSM in a system-identification task.

\vspace{-2pt}
\section{Preliminaries}
\label{sec:preliminaries}
\vspace{-6pt}
\subsection{Notation}
Bold uppercase and lowercase letters denote matrices and vectors, respectively. 
$\bI$ is the identity matrix.
We denote by $\mathbb{S}_{++}^n$ ($\mathbb{S}_{+}^n$) the set of $n \times n$ symmetric positive (semi-)definite matrices, $\mathbb{A}^n$ the $n \times n$ skew-symmetric matrices, and $\mathbb{D}_{++}^n$ the diagonal matrices with strictly positive entries. For a sequence $\mathbf{u} = (\mathbf{u}_t)_{t=0}^{T-1}$ and $\bQ \in \mathbb{S}_{++}^n$, we define the weighted norm $\Vert{}\mathbf{u}\Vert{}_\bQ^2 = \sum_{t=0}^{T-1} \mathbf{u}_t^\top \bQ \mathbf{u}_t$.
Where clear from context, the layer index $(l)$ (e.g., as in \eqref{eq:NN}) is omitted for brevity.

\vspace{-4pt}
\subsection{State-space network}
We consider an $L$-layer DNN \eqref{eq:NN} as depicted in Fig.\ \ref{fig:concept}.
Each layer $\LL^{(l)}$ is formulated as a state-space system with matching input and output dimensions, integrated with an element-wise activation function $\sigma(\cdot)$ whose slope is restricted to $[0,1]$ (e.g., ReLU, Sigmoid).
For $l = 1, \dots, L$, the discrete-time dynamics of the $l$-th layer $\vecy_t^{(l)}=\LL^{(l)}(\vecu_t^{(l)})$ $(t=0,\ldots,T-1)$ are governed by
\begin{equation}
\begin{aligned}
\vecx_{t+1}^{(l)} &= \bA^{(l)}\vecx_t^{(l)} + \bB^{(l)}\vecu_t^{(l)},\\
\vecy_t^{(l)} &= \sigma(\bC^{(l)}\vecx_t^{(l)} + \bD^{(l)}\vecu_t^{(l)}),
\end{aligned} \label{eq:ssm_activation}
\end{equation}
where $\vecx_t^{(l)} \in \mathbb{R}^{n_l}$, $\vecu_t^{(l)} \in \mathbb{R}^{m}$, and $\vecy_t^{(l)} \in \mathbb{R}^{m}$ represent the state, input, and output vectors at time step $t$, respectively. 
The system matrices $\bA^{(l)}\in\mathbb{R}^{n_l \times n_l}, \bB^{(l)}\in\mathbb{R}^{n_l \times m}, \bC^{(l)}\in\mathbb{R}^{m \times n_l}, \bD^{(l)}\in\mathbb{R}^{m\times m}$ are constructed from learnable parameters.
The output of each layer $\vecy_t^{(l)}$ serves as the input $\vecu_t^{(l+1)}$ to the subsequent layer.
The full-network input and output sequences are denoted by
$\vecu^{\mathrm{in}} = (\vecu_t^{(1)})_{t=0}^{T-1}$ and $\vecy^{\textrm{out}} = \NN(\vecu^{\mathrm{in}}) = (\NN(\vecu_t^{(1)}))_{t=0}^{T-1}$, respectively.

\subsection{Lipschitz continuity with weighted norms}\label{sec:lip}
The goal of this paper is to bounds the input-output sensitivity of the entire network, through a generalized notion of Lipschitz continuity.
\begin{dfn}
\label{dfn:lip}
Let $\NN$ denote an operator mapping $\vecu^{\mathrm{in}}$ to $\vecy^{\mathrm{out}}$. 
For positive-definite matrices $\bQ_{\rm in} \in \mathbb{S}_{++}^{m}$ and $\bQ_{\rm out} \in \mathbb{S}_{++}^{m}$, $\NN$ is 
$(\bQ_{\rm in},\bQ_{\rm out})$-Lipschitz continuous if 
\begin{equation}
    \bigl\| \NN(\vecu^{\mathrm{in}}) - \NN(\widetilde{\vecu}^{\mathrm{in}}) \bigr\|_{\bQ_{\rm out}} \le \bigl\| \vecu^{\mathrm{in}} - \widetilde{\vecu}^{\mathrm{in}} \bigr\|_{\bQ_{\rm in}} 
    \label{eq:lip}
\end{equation}
holds for any pair of admissible inputs $\vecu^{\mathrm{in}}$ and $\widetilde{\vecu}^{\mathrm{in}}$.
\end{dfn}
This weighted formulation accommodates varying sensitivities across layers, offering the flexibility to tailor robustness certificates without over-constraining the entire model.
\begin{remark}\label{rem}
Definition \ref{dfn:lip} recovers standard $\rho$-Lipschitz continuity as a special case under $\bQ_{\rm in}=\rho^2\bI$ and $\bQ_{\rm out}=\bI$. 
\end{remark}

\subsection{Sufficient condition for Lipschitz continuity of SSM}
\label{sec:dissipativity}
To construct neural architectures with certifiable Lipschitz bounds, recent studies  \cite{REN24,massai24interconnect_dissipative,LipKernel} leverage the theory of incremental dissipativity \cite{koelwijin21,verhoek23}. 
Building upon this theoretical foundation,
LipKernel \cite{LipKernel} established a layer-wise sufficient condition ensuring that the entire cascaded network $\mathcal{N}$ satisfies \eqref{eq:lip}. 
The following Theorem constitutes the core of this formulation, which directly follows from Theorem 7 and Lemma 8 of \cite{LipKernel}.

\begin{thm}
\label{thm:global_lipschitz}
Consider the system~\eqref{eq:ssm_activation}. 
Suppose there exist matrices $\bP\in\mathbb{S}_{++}^{n_l}$, $\bQ_{\mathrm{prev}}\in\mathbb{S}_{++}^{m}$, $\bQ\in\mathbb{S}_{++}^{m}$, $\bV\in\mathbb{D}_{++}^{m}$, such that 
\begin{equation}
\bS = 
\begin{bmatrix} \bP-\bA^\transp \bP \bA & -\bA^\transp \bP \bB & -\bC^\transp\bV \\ -\bB^\transp \bP \bA & \bQ_{\mathrm{prev}}-\bB^\transp \bP \bB & -\bD^\transp\bV \\ -\bV \bC & -\bV \bD & 2\bV-\bQ \end{bmatrix} \succeq 0
\label{eq:lmi_activation} 
\end{equation} 
for each layer $\LL^{(l)}~(l=1,\ldots,L)$.
Setting $\bQ_{\mathrm{prev}}^{(l)} = \bQ^{(l-1)}$, $\bQ^{(0)} = \bQ_{\mathrm{in}}$, and $\bQ^{(L)} = \bQ_{\mathrm{out}}$, the network $\NN$ is $(\bQ_{\mathrm{in}},\bQ_{\mathrm{out}})$-Lipschitz continuous under the condition $\vecx_{0}^{(l)} = \mathbf{0}$.
\end{thm}

To fulfill the requirements of Theorem \ref{thm:global_lipschitz}, we parametrize the matrices
$\{\bA^{(l)}, \bB^{(l)}, \bC^{(l)}, \bD^{(l)}, \bP^{(l)}, \bQ^{(l)}, \bV^{(l)}\}$ that satisfy \eqref{eq:lmi_activation}. 

\section{Proposed Architecture: LipSSM}
\label{sec:proposed_method} 
Building on Theorem \ref{thm:global_lipschitz}, we propose a DNN architecture termed LipSSM, constructed by cascading SSM layers whose system matrices are specifically parameterized to enforce a Lipschitz bound. 
To guarantee global Lipschitz continuity under a prescribed bound, these system matrices are computed as described in Section \ref{sec:learnable_and_system}, while the underlying theoretical rationale is detailed in Section \ref{sec:derivation}.

\subsection{Learnable parameters and construction of system matrices}\label{sec:learnable_and_system}
\begin{figure}
    \centering
    \includegraphics[width=0.99\linewidth]{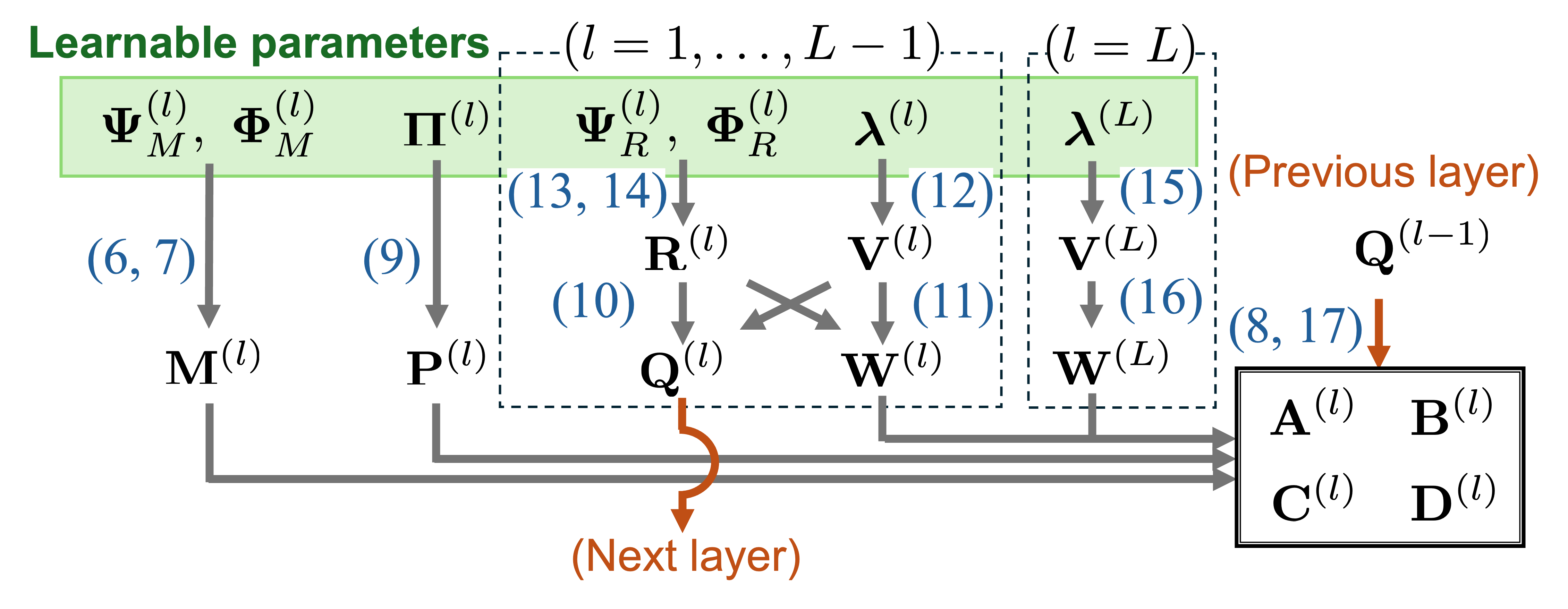}
    \vspace{-14pt}
    \caption{Learnable parameters and their relationships used to construct the system matrices for the $l$-th layer. 
    Numbers shown in blue correspond to equation numbers.}
    \label{fig:construction}
\end{figure}

LipSSM is an $L$-layer DNN as in \eqref{eq:NN}, where each layer is an SSM defined in \eqref{eq:ssm_activation}.
The system matrices $\{\bA^{(l)},\bB^{(l)}, \bC^{(l)}, \bD^{(l)}\}$ in \eqref{eq:ssm_activation} are specifically constructed to satisfy the requirements of Theorem \ref{thm:global_lipschitz}.
To this end, we define the learnable parameters as
\begin{equation}
    \{ \bPsi_M^{(l)}, \bPhi_M^{(l)}, \bPi^{(l)},\bPsi_R^{(l)}, \bPhi_R^{(l)}, \bm{\lambda}^{(l)} \}\quad \text{for}\quad l\in\{1,\ldots,L\},
    \label{eq:learnableParametersInLipsSSM}
\end{equation}
where $\bPsi_M^{(l)}\in\mathbb{R}^{(m+n_l)\times (m+n_l)}, \bPhi_M^{(l)}\in\mathbb{R}^{(m+n_l)\times (m+n_l)}$, $\bPi^{(l)}\in\mathbb{R}^{n_l\times n_l}$, $\bPsi_R^{(l)}\in\mathbb{R}^{m\times m}, \bPhi_R^{(l)}\in\mathbb{R}^{m\times m}$, $\bm{\lambda}^{\!(l)}\!\!\in\mathbb{R}^{m}$,
and compute the system matrices $\{\bA^{(l)},\bB^{(l)}$, $\bC^{(l)}, \bD^{(l)}\}$ from these learnable parameters. 
The step-by-step computational procedure is detailed below, with an overview depicted in Fig.~\ref{fig:construction}.

\subsubsection{Computation of normalized system matrices}\label{sec:construction_normalized}
To compute the system matrices of LipSSM, we first compute their normalized versions 
$\widetilde{\bA}, \widetilde{\bB}, \widetilde{\bC}$, and $\widetilde{\bD}$. 
Inspired by \cite{helfrich2018}, we apply the generalized Cayley transform as follows:
\begin{equation}
\bM = (\bI + \bK_M + \bY_M)^{-1} (\bI - \bK_M - \bY_M),
\label{eq:M_cayley}
\end{equation}
where $\bK_M$ is a skew-symmetric matrix, and $\bY_M$ is a positive semi-definite matrix, constructed using the learnable matrices as follows:
\begin{equation}
\bK_M = \bPsi_{M} - \bPsi_{M}^\top, \qquad\quad \bY_M = \bPhi_{M} \bPhi_{M}^\top.
\label{eq:first_step}
\end{equation}
Then, we partition the obtained matrix $\bM$ into four matrices as 
\begin{equation}
\bM =
\begin{bmatrix}
\widetilde{\bA} & \widetilde{\bB}\\ \widetilde{\bC} & \widetilde{\bD}
\end{bmatrix}.
\label{eq:M}
\end{equation}
These system matrices cannot take into account the information on the other layers, and therefore they are subsequently transformed using the weight matrices computed as follows.

\vspace{-2pt}
\subsubsection{Computation of weight matrices}\label{sec:construction_weight}
Matrices for transforming $\{\widetilde{\bA},~ \widetilde{\bB},~\widetilde{\bC},~ \widetilde{\bD}\}$ in \eqref{eq:M} are computed as follows.
First, positive-definite matrix $\bP$ is constructed from $\bPi$ as
\begin{equation}
    \bP^{(l)} = \bPi^{(l)} (\bPi^{(l)})^\transp + \epsilon \bI,
    \label{eq:P}
\end{equation}
where $\epsilon > 0$ is a small constant to ensure strict positivity. This matrix has no further structural constraint.

Next, $\bQ^{(l)}$ and $\bW^{(l)}$ are constructed from 
$\bm{\lambda}^{(l)}$, $\bPsi_R^{(l)}$, $\bPhi_R^{(l)}$ as 
\begin{align}
\bQ^{(l)} &= (\bV^{(l)})^{1/2} (\bR^{(l)})^\transp \bR^{(l)} (\bV^{(l)})^{1/2},\label{eq:X_med}\\
\bW^{(l)} &= (\bV^{(l)})^{1/2} \left( 2\bI - (\bR^{(l)})^\transp \bR^{(l)} \right)^{-1} (\bV^{(l)})^{1/2},\label{eq:W_med}
\end{align}
where $\bV^{(l)}$ and $\bR^{(l)}$ are computed as
\begin{align}
\bV^{(l)} &= \text{diag}(\text{softplus}(\bm{\lambda}^{(l)})),\label{eq:V_med} \\
\bR^{(l)} &= (\bI+\bK^{(l)}_R+\bY^{(l)}_R)^{-1} (\bI-\bK^{(l)}_R-\bY^{(l)}_R), \label{eq:R}\\
\bK_R^{(l)} &= \bPsi_{R}^{(l)} - (\bPsi_{R}^{(l)})^\transp,\qquad \bY_R^{(l)} = \bPhi_{R}^{(l)}(\bPhi_{R}^{(l)})^\transp
    \label{eq:KRYR}
\end{align}
These weight matrices are utilized in the intermediate layers (i.e., for $l = 1, \ldots, L-1$).

The final layer $\LL^{(L)}$ is treated differently.
For the final layer, 
$\bQ^{(L)}$ is set to the prescribed $\bQ_{\text{out}}$. 
The matrix $\bV^{(L)}$ is computed as
\begin{equation}
\bV^{(L)} = \frac{1}{2} \bar{\bQ} + \operatorname{diag}\left(\operatorname{softplus}(\bm{\lambda}^{(L)})\right),
\label{eq:V_term}
\end{equation}
where $\bar{\bQ} = \bQ_{\text{out}}$ if $\bQ_{\text{out}}$ is diagonal, and $\bar{\bQ} = \bigl\|\bQ_{\text{out}}\bigr\|_{2\,}\bI$ otherwise, with $\|\cdot\|_{2}$ denoting the spectral norm. $\bW^{(L)}$ is computed as
\begin{equation}
\bW^{(L)} = \bV^{(L)} (2\bV^{(L)} - \bQ_{\text{out}})^{-1} \bV^{(L)}.
\label{eq:W_term}
\end{equation}
Using these weights, the matrices in \eqref{eq:M} are transformed to the system matrices of the SSM layers.

\vspace{-2pt}
\subsubsection{Computation of system matrices}
Using $\bP$, $\bQ$, $\bW$ given in Section \ref{sec:construction_weight}, the system matrices are computed from $\{\widetilde{\bA},~ \widetilde{\bB},~\widetilde{\bC},~ \widetilde{\bD}\}$ in Section \ref{sec:construction_normalized} as follows:
\begin{equation} 
\label{eq:system_matrices} 
\begin{aligned} 
\bA &= \bP^{-1/2}\widetilde\bA\bP^{1/2},& \quad\;
\bB &= \bP^{-1/2}\widetilde\bB\bQ_{\mathrm{prev}}^{1/2},\\ 
\bC &= \bW^{-1/2}\widetilde\bC\bP^{1/2},& \quad\;
\bD &= \bW^{-1/2}\widetilde\bD\bQ_{\mathrm{prev}}^{1/2},
\end{aligned} \end{equation}
where $\bQ_{\mathrm{prev}}$ denotes the matrix $\bQ$ inherited from the preceding layer. 
These system matrices constitute the proposed LipSSM architecture, with the corresponding theoretical justification provided below.

\subsection{Main theoretical result: Lipschitz continuity of LipSSM}
\label{sec:main_theorem}
The proposed parameterization structurally fulfills the requirements of Theorem~\ref{thm:global_lipschitz}, guaranteeing Lipschitz continuity without any additional technique. 
This core result is formally stated as follows. 

\begin{thm}
\label{thm:main_lipssm_guarantee}
Let a DNN $\NN$ be a cascade of $L$ SSM layers defined in \eqref{eq:ssm_activation}, where each layer's system matrices $(\bA^{(l)}, \bB^{(l)}, \bC^{(l)}, \bD^{(l)})$ are constructed from learnable parameters in \eqref{eq:learnableParametersInLipsSSM} $(\bPi^{(l)}$, $\bPsi_M^{(l)}$, $\bPhi_M^{(l)}$, $\bPsi_R^{(l)}$, $\bPhi_R^{(l)}$, $\bm{\lambda}^{(l)})$  via \eqref{eq:M_cayley}--\eqref{eq:system_matrices} for all $l\in\{1,\ldots,L\}$.
Then, the DNN $\NN$ is guaranteed to be $(\bQ_{\mathrm{in}},\bQ_{\mathrm{out}})$-Lipschitz continuous.
\end{thm}

Due to space limitation, all the proofs are omitted. Theorem \ref{thm:main_lipssm_guarantee} follows from the derivations presented in the next section.

\section{Derivation of LipSSM parameterization}\label{sec:derivation}
This section presents the theoretical derivation of the parameterization of LipSSM proposed in Section \ref{sec:proposed_method}.
To fulfill the requirements of Theorem \ref{thm:global_lipschitz}, the matrices $\bA^{(l)}, \bB^{(l)}, \bC^{(l)}, \bD^{(l)}, \bP^{(l)}, \bQ^{(l)}, \bV^{(l)}$ require a parameterization scheme that inherently guarantees \eqref{eq:lmi_activation}.
In this section, we establish that the layer-wise condition \eqref{eq:lmi_activation} ($\bS \succeq 0$) holds if and only if the spectral norm condition $\Vert{}\bM\Vert{}_2 \le 1$ for $\bM$ in \eqref{eq:M} and the condition $2\bV - \bQ \succeq 0$ are satisfied.

Let the matrix $\bS$ in \eqref{eq:lmi_activation} be partitioned as follows:
\begin{equation*}
\bS = \begin{bmatrix} \bS_{11} & \bS_{12} \\ \bS_{12}^\transp & \bS_{22} \end{bmatrix},\quad
\bS_{11} = 
\begin{bmatrix} \bP-\bA^\transp \bP \bA & -\bA^\transp \bP \bB\\
-\bB^\transp \bP \bA & \bQ_{\mathrm{prev}}-\bB^\transp \bP \bB
\end{bmatrix} ,
\end{equation*}
with $\bS_{22} = 2\bV - \bQ$, and $\bS_{12} = -\begin{bmatrix} \bV\bC ~~ \bV\bD \end{bmatrix}^\transp$. 
Then, the condition $\bS \succeq 0$ in \eqref{eq:lmi_activation} holds if and only if
\begin{align}
\bS_{22} = 2\bV - \bQ \succeq 0, \label{eq:schur_cond}
\end{align}
and the Schur complement satisfies
\begin{align}
\bS_{11} - \bS_{12} \bS_{22}^{-1} \bS_{12}^\top = \bS_{11} - \begin{bmatrix} \bC ~~ \bD \end{bmatrix}^\top \bW \begin{bmatrix} \bC ~~ \bD \end{bmatrix} \succeq 0,
\label{eq:schur_complement_positivity}
\end{align}
where $\bW$ absorbs $\bV$ in $\bS_{12}$ as
\begin{equation}
\bW = \bV (2\bV - \bQ)^{-1} \bV. \label{eq:effective_metric}
\end{equation}
These two conditions \eqref{eq:schur_cond} and \eqref{eq:schur_complement_positivity} lead to the proposed parametrization in Section \ref{sec:proposed_method} as follows.

\vspace{-2pt}
\subsection{Derivation of system matrices} \label{sec:derivation_system}
Direct calculation shows that the condition \eqref{eq:schur_complement_positivity} is equivalent to
\begin{equation}\! \!\!\! \!\!\!
\begin{bmatrix}
\bA^\transp \bP\bA - \bP + \bC^\transp \bW\bC & \bA^\transp \bP\bB + \bC^\transp \bW\bD \\
\bB^\transp \bP\bA + \bD^\transp \bW\bC & \bB^\transp \bP\bB + \bD^\transp \bW\bD - \bQ_{\mathrm{prev}}
\end{bmatrix} \preceq 0.
\label{eq:lmi_reduced}
\end{equation}
Comparing this condition with \eqref{eq:M} and \eqref{eq:system_matrices} and using the properties of positive definite matrices, we obtain the following equivalent condition in terms of the spectral norm constraint.

\begin{lemma} \label{lem:contractive_equiv}
Let $\widetilde\bA, \widetilde\bB, \widetilde\bC, \widetilde\bD$, and $\bM$ be as in \eqref{eq:M}, where the relation to the system matrices is given by \eqref{eq:system_matrices} as $\widetilde\bA = \bP^{1/2}\bA\bP^{-1/2}$, $\widetilde\bB = \bP^{1/2}\bB\bQ_{\mathrm{prev}}^{-1/2}$,
$\widetilde\bC = \bW^{1/2}\bC\bP^{-1/2}$, and
$\widetilde\bD = \bW^{1/2}\bD\bQ_{\mathrm{prev}}^{-1/2}$.
Then, the condition \eqref{eq:lmi_reduced} holds if and only if $\bM^\transp \bM \preceq \bI$, which is equivalent to the spectral norm constraint $\Vert{}\bM\Vert{}_2 \le 1$.
\end{lemma}

To structurally enforce $\Vert{}\bM\Vert{}_2 \le 1$, we adopt a natural generalization of the Cayley transform, building upon the standard formulation utilized in \cite{helfrich2018}. 
The associated matrix properties are formalized in the following Proposition, which justifies \eqref{eq:M_cayley}. 

\begin{proposition}
\label{prop:cayley_parameterization}
Let $\bK_G \in \mathbb{A}^{n}$ and $\bY_G \in \mathbb{S}_{+}^{n}$ be arbitrary. 
Then, the generalized Cayley transform defined by
\begin{equation}
    \mathbf{G} = (\bI + \bK_G + \bY_G)^{-1} (\bI - (\bK_G + \bY_G))
    \label{eq:generalized_cayley}
\end{equation}
unconditionally satisfies $\|\mathbf{G} \|_2 \le 1$.
\end{proposition}

Using this transform in \eqref{eq:M_cayley}, we obtain $\bM$ satisfying $\|\bM\|_2\leq1$, and hence we call $\widetilde\bA, \widetilde\bB, \widetilde\bC, \widetilde\bD$ as normalized system matrices. Since these matrices cannot take into account information from the other layers, they are combined with weight matrices to obtain the system matrices in \eqref{eq:system_matrices}.

\vspace{-2pt}
\subsection{Derivation of weight matrices}\label{sec:derivation_metric}
For intermediate layers ($l = 1, \dots, L-1$), 
we set $\bV^{(l)}$ independently of the other variables, directly yielding \eqref{eq:V_med}.
We then construct $\bQ^{(l)}$ to fulfill the condition \eqref{eq:schur_cond}.
Factoring $2\bV^{(l)} - \bQ^{(l)}$ as 
\begin{equation}
    2\bV^{(l)} - \bQ^{(l)} = (\bV^{(l)})^{1/2} \big( 2\bI - (\bR^{(l)})^\transp \bR^{(l)} \big) (\bV^{(l)})^{1/2}
    \label{eq:aaa}
\end{equation}
leads to a condition equivalent to \eqref{eq:schur_cond} because $\bV\in\mathbb{D}_{++}^{m}$:
\begin{equation}
    2\bI - (\bR^{(l)})^\transp \bR^{(l)} \succeq 0.
\end{equation}

To guarantee this inequality, we generate $\bR^{(l)}$ using the generalized Cayley transform of a positive definite matrix $\bY^{(l)}_R$ and an anti-symmetric matrix $\bK^{(l)}_R$ as in \eqref{eq:KRYR} and \eqref{eq:R}. 
This construction of $\bR^{(l)}$ guarantees $(\bR^{(l)})^\transp \bR^{(l)} \preceq \bI$ by Proposition \ref{prop:cayley_parameterization} (as $\|\bR^{(l)}\|_2 \le 1$), thereby ensuring $2\bI - (\bR^{(l)})^\transp \bR^{(l)}\succ 0$.
Consequently, the parameterizations for $\bQ^{(l)}$ in \eqref{eq:X_med} is justified via \eqref{eq:aaa}.
Substituting \eqref{eq:V_med} into \eqref{eq:effective_metric} yields \eqref{eq:W_med}.

For the terminal layer ($l = L$),
to structurally enforce \eqref{eq:schur_cond} 
for a given $\bQ^{(L)}$, we parameterize $\bV^{(L)}$ using a learnable vector $\bm{\lambda}^{(L)}$ as defined in \eqref{eq:V_term}.
Since $\bQ^{(L)}=\bQ_{\text{out}} \preceq \bar{\bQ}$, this construction guarantees \eqref{eq:schur_cond}. 
Substituting \eqref{eq:V_term} into \eqref{eq:effective_metric} yields \eqref{eq:W_term}.

\section{Numerical Experiments}

\subsection{Numerical verification of the Lipschitz boundness}

This section numerically verifies that the Lipschitz constant of the proposed LipSSM, denoted by $\NN_\varTheta$, satisfies the prescribed Lipschitz-bound as we showed in Theorem~\ref{thm:main_lipssm_guarantee}.
Here, $\varTheta$ denotes the set of all learnable parameters in LipSSM, i.e., those listed in \eqref{eq:learnableParametersInLipsSSM}.

In this experiment, we consider the case when $\bQ_{\mathrm{in}}=\bQ_{\mathrm{out}}=\bI$, in which Theorem~\ref{thm:main_lipssm_guarantee} guarantees
$\Lip(\NN_\varTheta)\leq 1$ (see Remark~\ref{rem}), regardless of the choice of $\varTheta$.
To numerically examine this inequality, we compute the largest spectral norm of the Jacobian, given by
$B=
\sup_{(\mathbf{u}^{\mathrm{in}},\varTheta)}
\|
\bJ_\varTheta(\mathbf{u}^{\mathrm{in}})\|_{2}$,
where $\bJ_\varTheta(\mathbf{u}^{\mathrm{in}})$ denotes the Jacobian matrix of
$\NN_\varTheta$ at the input $\mathbf{u}^{\mathrm{in}}$.
From Theorem~\ref{thm:main_lipssm_guarantee}, $B\leq 1$ must hold because
$\|
\bJ_\varTheta(\mathbf{u}^{\mathrm{in}})\|_2\leq
\Lip(\NN_\varTheta)\leq 1$ for any $\mathbf{u}^{\mathrm{in}}$ and $\varTheta$.

Following the numerical validation strategy in~\cite{lipsam_icassp,lipsam26full},
we searched for 
$\sup_{(\mathbf{u}^{\mathrm{in}},\varTheta)}
\|
\bJ_\varTheta(\mathbf{u}^{\mathrm{in}})\|_{2}$
using the Adam optimizer.
During optimization, the spectral norm was approximated by five power iterations, allowing its gradients with respect to $\mathbf{u}^{\mathrm{in}}$ and $\varTheta$ to be computed via automatic differentiation.
The learning rate of Adam was set to $10^{-2}$, and the optimization was performed for $100$ iterations.
After optimization, the spectral norm at the obtained
$\mathbf{u}^{\mathrm{in}}$ and $\varTheta$ was evaluated exactly using singular value decomposition.
We tested 
network depths 
$L\in\{1,2,4,8\}$ and 
state dimensions
$n_l\in\{4,8,16,32\}$ $(l=1,\ldots,L)$ with
sequence length 
$T=32$.
For each configuration, we performed $100$ trials with random initializations of
$\mathbf{u}^{\mathrm{in}}$ and $\varTheta$.

Table~\ref{tbl:lips_attack} summarizes the results.
For every tested configuration, the value $B$ (i.e., the empirically evaluated worst-case Lipschitz constant of LipSSM) remained below $1$, which is consistent with Theorem~\ref{thm:main_lipssm_guarantee}.
Moreover, the obtained values of $B$ were consistently close to $1$, indicating that the proposed parameterization provides tight bounds without margin.

\subsection{IIR system identification} 
To evaluate LipSSM and LipKernel on sequence modeling, we conducted a single-channel nonlinear infinite impulse response (IIR) system identification benchmark defined by
\begin{equation*}
x_t = \alpha x_{t-1} + u_t, \quad y_t = \tanh(x_t), \quad t=1,\ldots,100,
\end{equation*}
with $x_0=0$, zero-mean Gaussian inputs $u_t$, and memory parameter $\alpha \in \{0.1, 0.5, 0.9\}$.

Both architectures consisted of two layers ($L=2$) with $\arctan(\cdot)$ activations. To cover the target system's gain across all $\alpha$ (requiring Lipschitz bounds of at least $1.1$, $2.0$, and $10$, respectively), the prescribed bound was set to $10$ for all $\alpha$ via $\bQ_{\mathrm{in}} = 10^2\bI$ and $\bQ_{\mathrm{out}}=\bI$. We set the state dimension of LipSSM to $2$, and evaluated LipKernel with kernel sizes $3$, $7$, and $11$.
Models were trained on $5000$ sequences and validated on $1000$ sequences using the Adam optimizer for $200$ epochs, selecting the checkpoint with the lowest validation loss. The empirical Lipschitz constant $\rho$ was evaluated using the maximum singular value of the network Jacobian. All experiments were conducted on an NVIDIA RTX 4090 GPU and an Intel Core Ultra 9 285K CPU. Training required approximately $10$ minutes for LipKernel and $100$ minutes for LipSSM.

\begin{table}[t]
\centering 
\caption{Empirical verification of the Lipschitz bound.} 
\vspace{2pt}
\label{tbl:lips_attack}
\scalebox{1.1}[1.05]{
\footnotesize{\begin{tabular}{c c c c c c} 
\toprule
\multirow{2}{*}{$L$} & \multirow{2}{*}{} & \multicolumn{4}{c}{State dimension ($n_l$)} \\
\cmidrule{3-6}
 & & 4 & 8 & 16 & 32 \\
\midrule
\multirow{3}{*}{1} 
 & Max  & 1.00000 & 1.00000 & 1.00000 & 0.99999 \\
 & Mean & 1.00000 & 1.00000 & 1.00000 & 0.99999 \\
 & Min  & 1.00000 & 1.00000 & 1.00000 & 0.99999 \\
\midrule
\multirow{3}{*}{2} 
 & Max  & 0.99999 & 0.99999 & 0.99996 & 0.99990 \\
 & Mean & 0.99999 & 0.99998 & 0.99995 & 0.99989 \\
 & Min  & 0.99998 & 0.99997 & 0.99995 & 0.99987 \\
\midrule
\multirow{3}{*}{4} 
 & Max  & 0.99990 & 0.99986 & 0.99969 & 0.99920 \\
 & Mean & 0.99989 & 0.99981 & 0.99960 & 0.99905 \\
 & Min  & 0.99988 & 0.99977 & 0.99952 & 0.99892 \\
\midrule
\multirow{3}{*}{8} 
 & Max  & 0.99946 & 0.99930 & 0.99883 & 0.99763 \\
 & Mean & 0.99935 & 0.99904 & 0.99859 & 0.99695 \\
 & Min  & 0.99928 & 0.99879 & 0.99809 & 0.99600 \\
\bottomrule
\end{tabular}}}
\end{table}
\begin{table}[t] 
\caption{Nonlinear IIR system identification performance.
Numbers in parentheses indicate the kernel size for LipKernel and the state dimension $n_1(=n_2)$ for LipSSM. 
} 
\vspace{2pt}
\label{tab:iir_comparison} 
\centering 
\scalebox{1.05}[1.05]{
\footnotesize{\begin{tabular}{c l r r r r r} \toprule 
$\alpha$ & Model & \multicolumn{1}{c}{NMSE} & Params & Infer. [ms] & \multicolumn{1}{c}{$\rho$} \\
\midrule
0.1 & LipKernel ($3$) & $0.2667$ & 22 & 0.420 & 0.9 \\ 
0.1 & LipKernel ($7$) & $0.1293$ & 94 & 0.432 & 0.7\\ 
0.1 & LipKernel ($11$) & $0.3085$ & 230  & 0.428 & 1.0\\ 
\rowcolor{highlightgray}
0.1 & LipSSM ($2$) & $\bm{0.0005}$ & 50 & 5.844 & 1.1 \\ 
\midrule
0.5 & LipKernel ($3$) & $0.2953$  & 22 & 0.446 & 1.2\\
0.5 & LipKernel ($7$) & $0.1754$ & 94 & 0.462 & 1.1\\ 
0.5 & LipKernel ($11$) & $0.3652$ & 230 & 0.471 & 1.5\\ 
\rowcolor{highlightgray}
0.5 & LipSSM ($2$) & $\bm{0.0006}$ & 50 & 3.583 &  1.9\\ 
\midrule
0.9 & LipKernel ($3$) & $0.7027$ & 22 & 0.425 & 1.1\\ 
0.9 & LipKernel ($7$) & $0.4304$ & 94 & 0.429 & 1.7\\ 
0.9 & LipKernel ($11$) & $0.5541$ & 230 & 0.465 & 1.8\\ 
\rowcolor{highlightgray}
0.9 & LipSSM ($2$) & $\bm{0.0031}$ & 50 & 5.034 & 6.0\\ 
\bottomrule 
\end{tabular} }}
\end{table}

Table~\ref{tab:iir_comparison} shows that LipSSM consistently outperforms LipKernel in normalized mean squared error (NMSE) across all $\alpha$, even in the short-memory regime ($\alpha = 0.1$), with a comparable or smaller model footprint. Furthermore, LipSSM achieves larger Lipschitz constants at higher $\alpha$, directly reflecting the theoretical gain $1/(1-\alpha)$ of IIR dynamics. Although LipKernel remains faster due to its efficient convolution, this highlights a trade-off between expressiveness and computational cost of Lipschitz-continuous DNNs.
\vspace{4pt}

\section{Conclusion}
This paper proposed LipSSM, a Lipschitz-continuous state-space architecture paired with an unconstrained parameterization that enforces prescribed Lipschitz bounds by construction.
Theoretical analysis together with numerical evaluations confirm the satisfaction of Lipschitz bounds and demonstrate the efficacy of LipSSM in nonlinear system identification. Future research will include scaling LipSSM to larger-scale sequence modeling tasks and extending the proposed parameterization to more expressive SSM.

\newpage
\bibliographystyle{IEEEtran}
\bibliography{refs}

@inproceedings{cisse2017,
  title = {Parseval Networks: {Improving} Robustness to Adversarial Examples},
  author = {Moustapha Cisse and Piotr Bojanowski and Edouard Grave and Yann Dauphin and Nicolas Usunier},
  booktitle = {Int. Conf. Mach. Learn. (ICML)},
  pages = {854--863},
  year = {2017},
  volume = {70},
}

@inproceedings{tsuzuku2018,
author = {Tsuzuku, Yusuke and Sato, Issei and Sugiyama, Masashi},
title = {Lipschitz-margin training: {Scalable} certification of perturbation invariance for deep neural networks},
year = {2018},
booktitle = {Adv. Neural Inf. Process. Syst. (NeurIPS)},
pages = {6542--6551},
numpages = {10},
}

@inproceedings{fazlyab19IQC,
 author = {Fazlyab, Mahyar and Robey, Alexander and Hassani, Hamed and Morari, Manfred and Pappas, George},
 booktitle = {Adv. Neural Inf. Process. Syst. (NeurIPS)},
 pages = {11427--11438},
 title = {Efficient and Accurate Estimation of {Lipschitz} Constants for Deep Neural Networks},
 volume = {32},
 year = {2019}
}

@INPROCEEDINGS{pnp2013,
  author={Venkatakrishnan, Singanallur V. and Bouman, Charles A. and Wohlberg, Brendt},
  booktitle={IEEE Glob. Conf. Signal Inf. Process.}, 
  title={Plug-and-Play priors for model based reconstruction}, 
  year={2013},
  volume={},
  number={},
  pages={945--948},}

@article{pnp2022zhang,
  author={Zhang, Kai and Li, Yawei and Zuo, Wangmeng and Zhang, Lei and Van Gool, Luc and Timofte, Radu},
  journal={IEEE Trans. Pattern Anal. Mach. Intell.}, 
  title={Plug-and-Play Image Restoration With Deep Denoiser Prior}, 
  year={2022},
  volume={44},
  number={10},
  pages={6360--6376},}

@INPROCEEDINGS{lipsam_icassp,
  author={Matsumoto, Kazuki and Uchida, Ren and Yatabe, Kohei},
  booktitle={IEEE Int. Conf. Acoust. Speech Signal Process. (ICASSP)}, 
  title={{LipsAM}: {Lipschitz}-Continuous Amplitude Modifier for Audio Signal Processing and its Application to Plug-And-Play Dereverberation}, 
  year={2026},
  volume={},
  number={},
  pages={816--820},
  }

@article{lipsam26full,
      title={{LipsAM}: {Lipschitz}-continuous Neural Networks for Convergent Plug-and-Play Audio Signal Recovery}, 
      author={ Matsumoto, Kazuki and Uchida, Ren and Yoshino, Natsuki and Yatabe, Kohei},
      journal = {ArXiv:2608.23038},
      year={2026},
      eprint={2608.23038},
      archivePrefix={arXiv},
      primaryClass={cs.SD},
      url={https://arxiv.org/abs/2608.23038}, 
}

@ARTICLE{pnp_converge2017,
  title = {Plug-and-Play {{ADMM}} for Image Restoration: {{Fixed}}-Point Convergence and Applications},
  author = {Chan, Stanley H. and Wang, Xiran and Elgendy, Omar A.},
  year = 2017,
  journal = {IEEE Trans. Comput. Imaging},
  volume = {3},
  number = {1},
  pages = {84--98},
}

@article{ulrike,
  title={Distance-based classification with {Lipschitz} functions},
  author={von Luxburg, Ulrike and Bousquet, Olivier},
  journal={J. Mach. Learn. Res.},
  volume={5},
  pages={669--695},
  year={2004}
}

@inproceedings{bartlett,
 author = {Bartlett, Peter and Foster, Dylan J and Telgarsky, Matus J},
 booktitle = {Adv. Neural Inf. Process. Syst. (NeurIPS)},
 pages = {6241--6250},
 title = {Spectrally-normalized margin bounds for neural networks},
 volume = {30},
 year = {2017}
}

@inproceedings{1LipLayers,
  author={Prach, Bernd and Brau, Fabio and Buttazzo, Giorgio and Lampert, Christoph H.},
  booktitle={IEEE/CVF Conf. Comput. Vis. Pattern Recognit. (CVPR)}, 
  title={{1-Lipschitz} Layers Compared: {Memory}, Speed, and Certifiable Robustness}, 
  year={2024},
  volume={},
  number={},
  pages={24574--24583},
}

@inproceedings{
trockman21orthogonal,
title={Orthogonalizing Convolutional Layers with the {Cayley} Transform},
author={Asher Trockman and J Zico Kolter},
booktitle={Int. Conf. Learn. Represent. (ICLR)},
year={2021},
}

@InProceedings{AOL22,
author={Prach, Bernd and Lampert, Christoph H.},
title={Almost-Orthogonal Layers for Efficient General-Purpose {Lipschitz} Networks},
booktitle={Eur. Conf. Comput. Vis. (ECCV)},
year={2022},
pages={350--365},
}

@inproceedings{meunier2022,
  title = 	 {A Dynamical System Perspective for {Lipschitz} Neural Networks},
  author =       {Meunier, Laurent and Delattre, Blaise J and Araujo, Alexandre and Allauzen, Alexandre},
  booktitle = 	 {Int. Conf. Mach. Learn. (ICML)},
  pages = 	 {15484--15500},
  year = 	 {2022},
  volume = 	 {162},
}

@inproceedings{wang23sandwich,
  title = 	 {Direct Parameterization of {Lipschitz}-Bounded Deep Networks},
  author =       {Wang, Ruigang and Manchester, Ian},
  booktitle = 	 {Int. Conf. Mach. Learn. (ICML)},
  pages = 	 {36093--36110},
  year = 	 {2023},
  volume = 	 {202},
}

@article{koelwijin21,
  author  = {Koelewijn, Patrick J. W. and T{\'o}th, Roland},
  title   = {Incremental stability and performance analysis of discrete-time nonlinear systems using the {LPV} framework},
  journal = {IFAC-PapersOnLine},
  volume  = {54},
  number  = {8},
  pages   = {75--82},
  year    = {2021},
}

@article{verhoek23,
  title   = {Convex incremental dissipativity analysis of nonlinear systems},
  author  = {Verhoek, Chris and Koelewijn, Patrick J. W. and Haesaert, Sofie and T{\'o}th, Roland},
  journal = {Automatica},
  volume  = {150},
  pages   = {110859},
  year    = {2023},
  doi     = {10.1016/j.automatica.2023.110859}
}

@ARTICLE{pauli2024a,
      title={Lipschitz constant estimation for general neural network architectures using control tools}, 
      author={Pauli, Patricia and Gramlich, Dennis and Allgöwer,  Frank},
      journal = {ArXiv:2405.01125},
      year={2024},
      eprint={2405.01125},
      archivePrefix={arXiv},
      primaryClass={cs.LG},
      url={https://arxiv.org/abs/2405.01125},    
}

@ARTICLE{rosser1975,
  author={Roesser, R.},
  journal={IEEE Trans. Autom. Control}, 
  title={A discrete state-space model for linear image processing}, 
  year={1975},
  volume={20},
  number={1},
  pages={1--10}}

@article{gramlich2026,
title = {Convolutional neural networks as {2-D} systems},
journal = {Automatica},
volume = {187},
pages = {112876},
year = {2026},
author = {Dennis Gramlich and Patricia Pauli and Carsten W. Scherer and Christian Ebenbauer},
}

@INPROCEEDINGS{pauli2023cayley,
  author={Pauli, Patricia and Wang, Ruigang and Manchester, Ian R. and Allgöwer, Frank},
  booktitle={IEEE Conf. Decis. Control (CDC)}, 
  title={{Lipschitz}-Bounded {1D} Convolutional Neural Networks using the {Cayley} Transform and the Controllability {Gramian}}, 
  year={2023},
  volume={},
  number={},
  pages={5345--5350},
}

@article{LipKernel,
title = {{LipKernel}: {Lipschitz}-bounded convolutional neural networks via dissipative layers},
journal = {Automatica},
volume = {188},
pages = {112959},
year = {2026},
author = {Patricia Pauli and Ruigang Wang and Ian R. Manchester and Frank Allgöwer},
}

@ARTICLE{REN24,
  author={Revay, Max and Wang, Ruigang and Manchester, Ian R.},
  journal={IEEE Trans. Autom. Control}, 
  title={Recurrent Equilibrium Networks: {Flexible} Dynamic Models With Guaranteed Stability and Robustness}, 
  year={2024},
  volume={69},
  number={5},
  pages={2855--2870},
}

@INPROCEEDINGS{massai24interconnect_dissipative,
  author={Massai, Leonardo and Saccani, Danilo and Furieri, Luca and Ferrari-Trecate, Giancarlo},
  booktitle={Eur. Control Conf. (ECC)}, 
  title={Unconstrained Learning of Networked Nonlinear Systems via Free Parametrization of Stable Interconnected Operators}, 
  year={2024},
  volume={},
  number={},
  pages={651--656},
  }

@InProceedings{helfrich2018,
  title = 	 {Orthogonal Recurrent Neural Networks with Scaled {Cayley} Transform},
  author =       {Helfrich, Kyle and Willmott, Devin and Ye, Qiang},
  booktitle = 	 {Int. Conf. Mach. Learn. (ICML)},
  pages = 	 {1969--1978},
  year = 	 {2018},
  volume = 	 {80},
}

@INPROCEEDINGS{L2RU25,
  author={Massai, Leonardo and Ferrari-Trecate, Giancarlo},
  booktitle={IEEE Conf. Decis. Control (CDC)}, 
  title={Free Parametrization of {L2}-bounded State Space Models}, 
  year={2025},
  volume={},
  number={},
  pages={7012--7017}
  }

@article{R2DN26,
      title={{R2DN}: {Scalable} Parameterization of Contracting and {Lipschitz} Recurrent Deep Networks}, 
      author={Nicholas H. Barbara and Ruigang Wang and Ian R. Manchester},
      journal = {ArXiv:2504.01250},
      year={2026},
      eprint={2504.01250},
      archivePrefix={arXiv},
      primaryClass={cs.LG},
      url={https://arxiv.org/abs/2504.01250}, 
}

@INPROCEEDINGS{martinelli23REN,
  author={Martinelli, Daniele and Galimberti, Clara Lucía and Manchester, Ian R. and Furieri, Luca and Ferrari-Trecate, Giancarlo},
  booktitle={IEEE Conf. Decis. Control (CDC)}, 
  title={Unconstrained Parametrization of Dissipative and Contracting Neural Ordinary Differential Equations}, 
  year={2023},
  volume={},
  number={},
  pages={3043--3048}
  }

@INPROCEEDINGS{fu2022hippo,
  title = {Hungry Hungry Hippos: {Towards} Language Modeling with State Space Models},
  author = {Daniel Y. Fu and Tri Dao and Khaled K. Saab and Armin W. Thomas and Atri Rudra and Christopher R{\'e}},
  booktitle = {Int. Conf. Learn. Represent. (ICLR)},
  year = {2023},
}

@inproceedings{
gu22,
title={Efficiently Modeling Long Sequences with Structured State Spaces},
author={Albert Gu and Karan Goel and Christopher Re},
booktitle = {Int. Conf. Learn. Represent. (ICLR)},
year={2022},
}

@article{gu24mamba,
      title={Mamba: {Linear}-Time Sequence Modeling with Selective State Spaces}, 
      author={Albert Gu and Tri Dao},
      journal = {ArXiv:2312.00752},
      year={2024},
      eprint={2312.00752},
      archivePrefix={arXiv},
      primaryClass={cs.LG},
      url={https://arxiv.org/abs/2312.00752}, 
}
\end{document}